\documentclass{article}
\usepackage{spconf,amsmath,graphicx, amsfonts}
\usepackage{comment, tabularx}

\usepackage{booktabs}
\usepackage{subfigure}
\ninept

\usepackage{xcolor}
\usepackage{cite}
\usepackage{xcolor}

\newcommand{\thanh}[1]{\textcolor{black}{#1}}
\newcommand{\kai}[1]{\textcolor{black}{#1}} 
\newcommand{\kaiw}[1]{\textcolor{black}{#1}} 
\title{A Neural Prosody Encoder For End-to-End Dialogue Act Classification}
\name{%
\begin{tabular}{@{}c@{}}
Kai Wei \textsuperscript{\rm 1}\textsuperscript{\rm *} \qquad 
Dillon Knox \textsuperscript{\rm 2}\thanks{\textsuperscript{\rm 2}Work done during author's internship at Amazon Alexa.}\textsuperscript{\rm *}\thanks{\textsuperscript{\rm *}Equal contribution.} \qquad 
Martin Radfar \textsuperscript{\rm 1} \qquad 
Thanh Tran \textsuperscript{\rm 1}  \qquad 
Markus M{ü}ller \textsuperscript{\rm 1 }  \qquad \\
\textit{Grant P. Strimel} \textsuperscript{\rm 1} \qquad
\textit{Nathan Susanj} \textsuperscript{\rm 1} \qquad
\textit{Athanasios Mouchtaris} \textsuperscript{\rm 1}  \qquad
\textit{Maurizio Omologo} \textsuperscript{\rm 1}  \qquad
\end{tabular}}

\address{ \textsuperscript{\rm 1} Alexa Speech, Amazon, \textsuperscript{\rm 2} University of Southern California}

\begin{document}
%
\maketitle

\begin{abstract}
Dialogue act classification (DAC) is a critical task for spoken language understanding in dialogue systems. Prosodic  \kai{features} such as \thanh{energy} and pitch \kaiw{have} been shown to be \kai{useful} for DAC. 
Despite \kaiw{their} importance, little research has explored neural approaches to integrate prosodic \kai{features} into end-to-end (E2E) DAC models \kaiw{which} infer dialogue acts directly from audio signals.
\kai{In this work, we propose an E2E neural architecture that takes into account this need of characterizing prosodic phenomena co-occurring at different levels inside an utterance.}
\kai{A novel part of this architecture is a learnable gating mechanism that assesses the importance of prosodic \kai{features} and selectively retains core information necessary for E2E DAC.}
\kai{Our proposed model improves the dialogue act  accuracy by 1.07\% absolute across three publicly available benchmark datasets.}

\end{abstract}
\section{Introduction}
Dialogue acts (DAs) are speech acts that represent intentions behind a user's request to achieve a conversational goal \cite{austin1962how}. Dialogue act classification (DAC) models aim to discriminate speech act units such as statement, question, backchannel, and agreement. For instance, when a user says ``yes'', DAC models \kai{are used to} determine whether the user's intent is to agree with what the voice assistant system has said (DA: agreement) or to signal that the user is paying attention to the system (DA: backchannel). 

\kai{Recent years have seen significant success in applying deep learning approaches to DAC \cite{tran2017a, ji2016a, ortega2018lexico, shen2016neural, dang2020endtoend, he2018exploring}. These approaches use either transcripts~\cite{tran2017a, ji2016a, shen2016neural} or a combination of transcript and audio~\cite{he2018exploring, ortega2018lexico, julia2010dialog} to predict DA. However, relying on transcripts has three limitations: First, transcripts are not always available for a spoken dialogue system. Second, collecting oracle transcripts is expensive. Third, errors introduced from transcribing audio have been shown to decrease the performance of DAC significantly~\cite{Tran_CNN}. More recently, \cite{dang2020endtoend} introduced an end-to-end (E2E) DAC approach, where DAs are directly inferred from audio signals. This approach can address the limitations of using transcripts as the inputs. Yet, how to effectively model audio signals for E2E DAC is underexplored. }

\kai{Prosody comprises the intonation, rhythm, and stress of spoken language. As highlighted in~\cite{wallbridge2021it}, it represents the non-lexical channel that serves a fundamental role in speech communication among humans. It captures the complex linguistic and semantic contents embedded in spoken language beyond words and their literal meanings \cite{dahan2015prosody}. At the syllable/word level, stressing on different syllables of a word can lead to different meanings (e.g., REcord vs. reCORD) \cite{ward2019prosodic}. At the sentence level, overall intonational contour contributes to characterize speaker’s intention and communicative meanings (e.g., agreement vs. backchannel: yes  vs. yes?)~\cite{honda2004physiological}. This highly intuitive linguistic phenomena inspired many works to explore ways to incorporate prosodic \kai{features} for DAC \cite{shriberg1998can, zimmermann2009joint, quarteroni2011simultaneous, ang2005automatic, stolcke2000dialogue, arsikere2016novel, Tran_CNN}.} Early research primarily focused on conventional cumulative-statistics~\cite{shriberg1998can} and traditional machine learning approaches~\cite {arsikere2016novel, zimmermann2009joint}. Of note, \cite{arsikere2016novel} found that the location of the maximum F0 occurrence can effectively distinguish between questions and statements.\cite{gravano2007on,benus2007the} also show that a pitch contour rises on the second syllable of words such as \textit{okay} can mark a topic shift as well as conveying affirmation or a backchannel~\cite{gravano2007on}. Recently, neural modeling has emerged as a promising yet understudied approach to encode prosodic \kai{features}. For instance, convolutional neural networks are used in \cite{Tran_CNN} \kai{to model sentence-level prosodic \kai{features}. However, little research has focused on neural approaches that fuse prosodic and spectral characteristics of audio signals at both syllable/word level and sentence level. Moreover, how best to integrate prosodic \kai{features} for E2E DAC remains unexplored.}

In this work, we propose a novel E2E neural architecture that take\kai{s} into account this need of characterizing prosodic phenomena co-occurring at different levels inside an utterance. An essential part of this architecture is a learnable gating mechanism that assesses the importance of prosodic \kai{features} and selectively retains core \kai{features} necessary for E2E DAC.
We compare our proposed model with previous E2E DAC models \cite{ortega2018lexico, he2018exploring} \kai{that only use spectral-based audio features}.  \kai{The results show that our models outperform the reference ones. Further,} we compare our neural prosody encoder with \kai{the state-of-the-art prosody neural encoder} \cite{Tran_CNN} on three public benchmark datasets: DSTC2~\cite{second}, and DSTC3~\cite{third}, and Switchboard~\cite{godfrey1992switchboard}. \kai{We show that} our proposed model outperforms \cite{Tran_CNN} on all these datasets. We also examine the effects of the gating mechanism and different prosodic features on our proposed model.

\section{Proposed Models}

\newcommand{\dialoglabels}{\{y^\text{diag.}\}}

\newcommand{\pitch}{\boldsymbol{c}}
\newcommand{\energy}{\boldsymbol{e}}
\newcommand{\prosodyframe}{\boldsymbol{p}}
\newcommand{\prosodylist}{P}
\newcommand{\prosodylistextend}{\prosodylist = \{(\energy_1, \pitch_1),...,(\energy_t, \pitch_t)\}}

\newcommand{\audioframe}{\boldsymbol{x}}
\newcommand{\audioframelist}{X=\{\audioframe_1, \audioframe_2, ..., \audioframe_t\}}

\newcommand{\lfbe}{\ell}
\newcommand{\lfbelist}{L=\{\lfbe_1, \lfbe_2, ..., \lfbe_t\}}

\newcommand{\localgate}{\beta}
\newcommand{\globalgate}{\gamma}
\newcommand{\globalprosodiclistname}{V}
\newcommand{\globalprosodic}{\boldsymbol{v}}

\newcommand{\acoustic}{\boldsymbol{a}}
\newcommand{\acousticlistname}{A}
\newcommand{\acousticlist}{\acousticlistname=\{\acoustic_1, \acoustic_2, ..., \acoustic_t\}}

\newcommand{\lstmhidden}{\boldsymbol{h}}
\newcommand{\lstmhiddenlistname}{H}
\newcommand{\lstmhiddenlist}{\lstmhiddenlistname=\{\lstmhidden_1^{(n)}, \lstmhidden_2^{(n)}, ..., \lstmhidden_t^{(n)}\}}

\newcommand{\similaritymatrix}{A^{(s)}}
\newcommand{\dissimilaritymatrix}{A^{(d)}}

\newcommand{\globalgatematrix}{G}

\newcommand{\globalfusion}{F}

\begin{figure}[htp]
    \centering
    \includegraphics[width=\columnwidth]{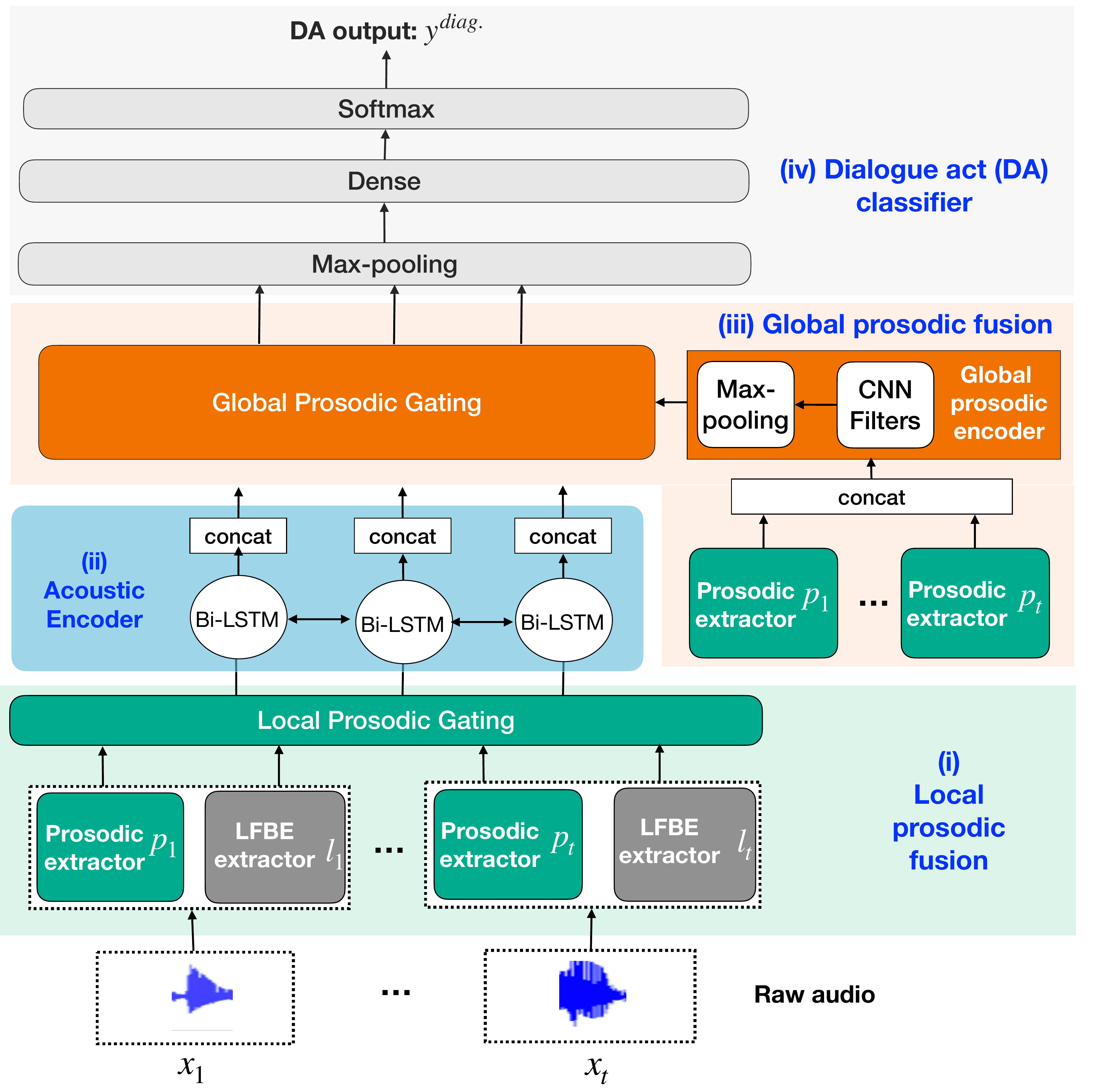}
    \vspace{-10pt}
    \caption{Our proposed E2E DAC model.}
    \label{fig:model}
    	\vspace{-10pt}
\end{figure}

We formulate the problem of E2E DAC tasks as follows: 
The input is a sequence of raw audio with $t$ time frames, 
$\audioframelist$. Each $x_t$ is converted to the logarithm of \kai{mel-scale filter bank} energy (LFBE) features 
$\lfbelist$ and prosodic features $\prosodylist = \{(\energy_1, \pitch_1),...,(\energy_t, \pitch_t)\}$,
where $\energy_i \in \mathcal{R}^{|\energy_i|}$ and $\pitch_i \in \mathcal{R}^{|\pitch_i|}$ denote energy and pitch features, respectively. 
Our goal is to correctly \kai{classify} DAs for each audio input $X$, namely $\dialoglabels$.  Figure~\ref{fig:model} shows our proposed model. It consists of (i) a 
{local prosodic infusion}, (ii) an {acoustic encoder}, (iii) a {global prosodic infusion}, and (iv) a {DA classifier}. We detail each component below.

\subsection{Local prosodic fusion}
The local prosodic infusion encodes prosodic features and infuses them with the LFBE features via our local prosodic gating. 
We extract the LFBE features using Kaldi \cite{kaldi} with a window size of 25 ms, a frame rate of 10 ms, and a sampling frequency of 8 kHz. We describe prosodic extractor and local prosodic gating below:

\subsubsection{Prosodic extractors}




 \noindent\textbf{Input:} We extract two types of \kai{basic} prosodic features: energy and pitch. We focus on these two features as they were found to be \kai{most important} for DAC~\cite{arsikere2016novel, Tran_CNN}.
\begin{itemize}
    \item Energy: For each audio frame $\audioframe_i \in X$, the 3-dimensional energy features $\energy_i$ are computed from the 40-mel frequency filter-bank using Kaldi \cite{kaldi}, in the same manner as \cite{Tran_CNN}. These features are (i) the log of total energy normalized by dividing by the maximum total energy of the utterance, (ii) the log of total energy in the lower 20 mel-frequency bands normalized by total energy, and (iii) the log of total energy in the higher 20 mel-frequency bands, normalized by total energy. 
    \item Pitch:  For each audio frame $\audioframe_i \in X$, the 3-dimensional pitch features $\pitch_i$ are (i) the warped Normalized Cross Correlation Function (NCCF), (ii) log-pitch with Probability of Voicing (POV)-weighted mean subtraction over a 1.5-second window, and (iii) the estimated derivative of the raw log pitch~\cite{ghahremani2014a}. 
   \end{itemize}



\noindent\textbf{Processing:} We first concatenate energy $\energy_i$ and pitch $\pitch_i$ for each audio frame $\audioframe_i \in X$. Then, we transform the concatenated  $\energy_i$ and $\pitch_i$ using the linear projection $W^{\energy\pitch}$ with the $ReLU$ activation function. 

\begin{equation}
	\label{eq:prosody}
	\prosodyframe_i = ReLU (W^{\energy\pitch} [\energy_i; \pitch_i]  
	)
\end{equation}
\noindent\textbf{Output:} We produce $\prosodylist=\{\prosodyframe_1, \prosodyframe_2,...,\prosodyframe_t\}$ as a stack of $t$ \kai{local prosodic embeddings} corresponding to $t$ audio frames of the input audio $X$, with each $\prosodyframe_i \in \prosodylist$ computed by Eq.~(\ref{eq:prosody}). 

\subsubsection{Local Prosodic Gating} 
\begin{figure}[htp]
	\centering
	\subfigure[Local Prosodic Gating.]{\includegraphics[width=0.46\linewidth]{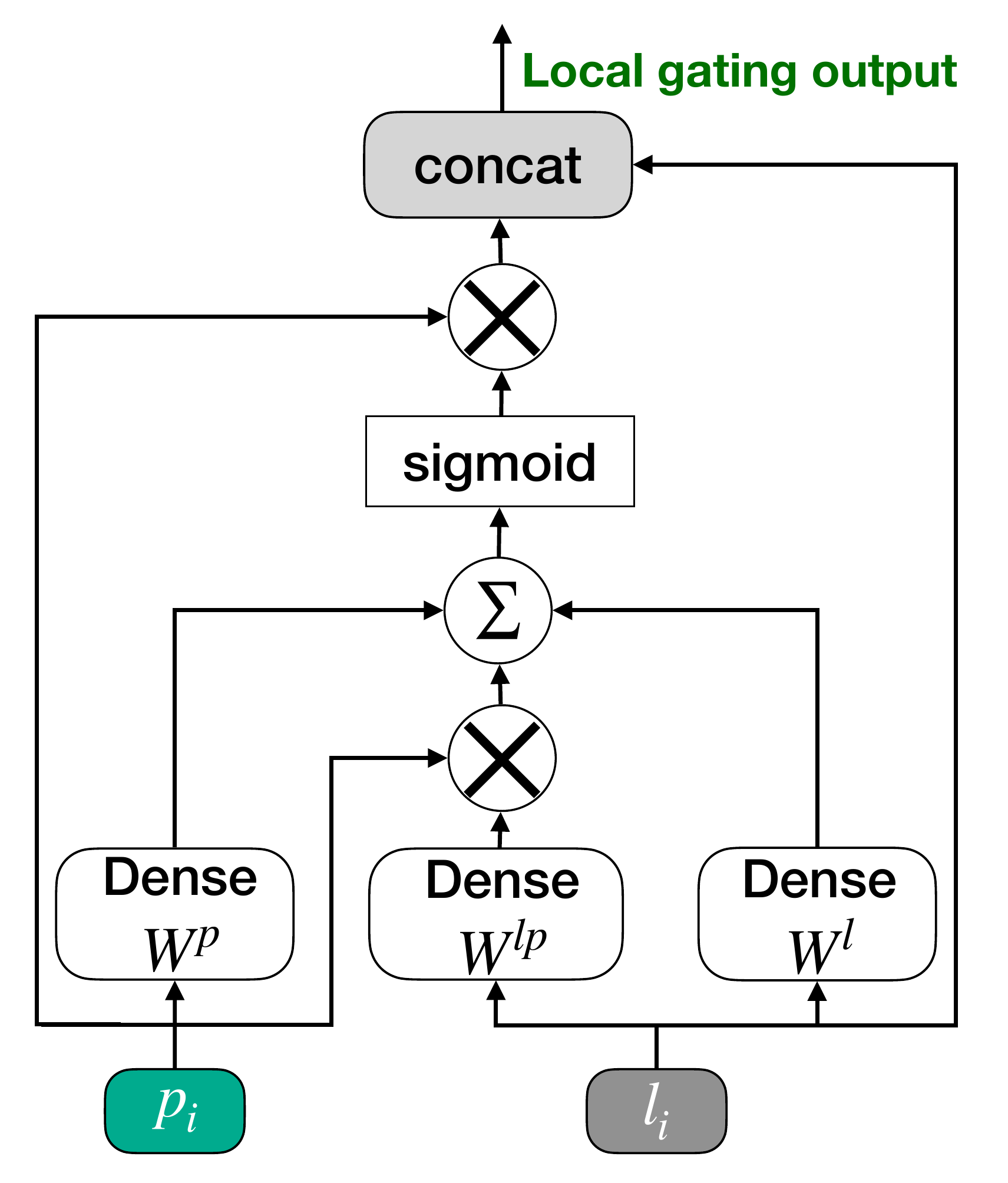}\label{fig:local-gate}}
	\subfigure[Global Prosodic Gating.]{\includegraphics[width=0.53\linewidth]{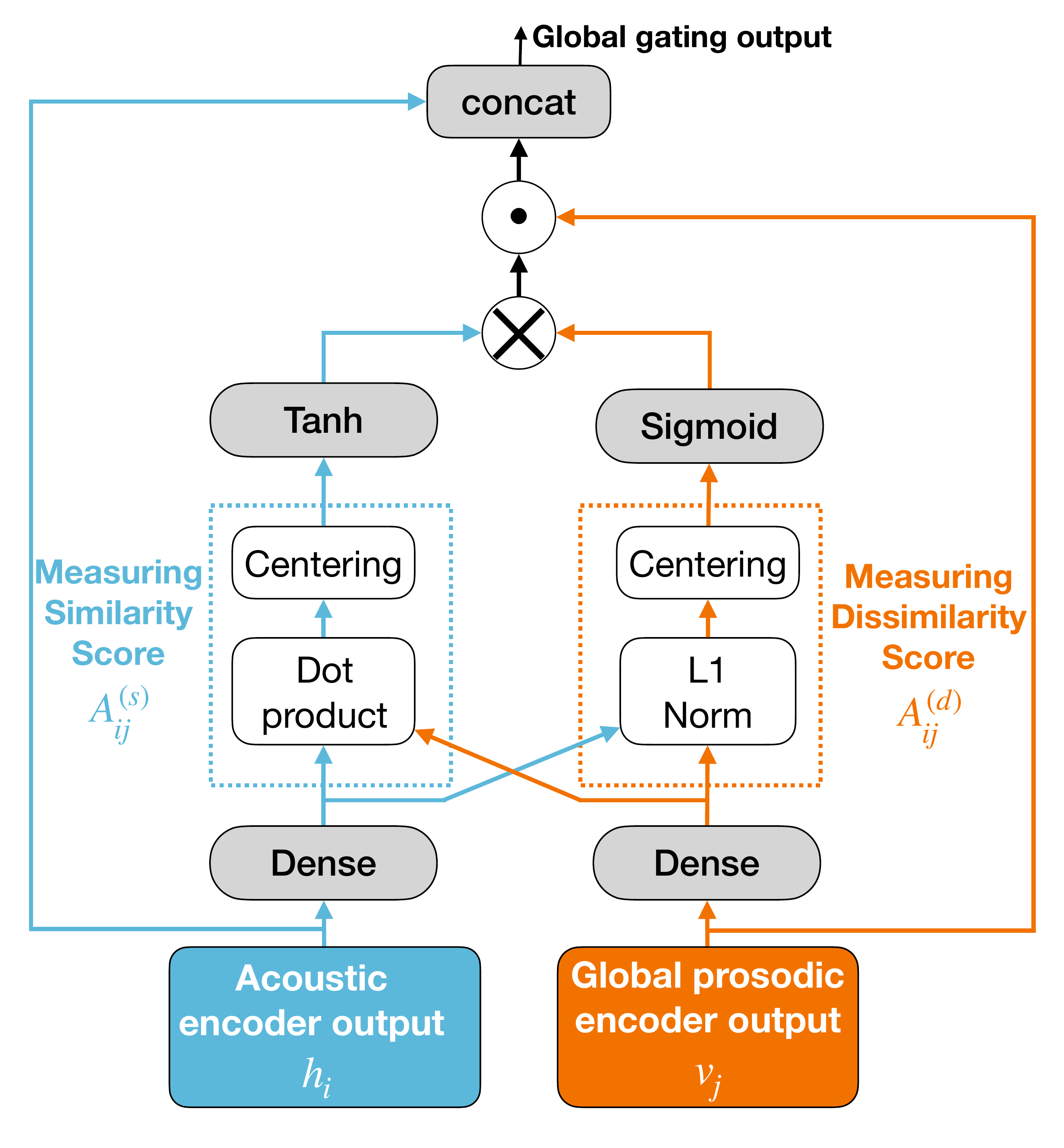}\label{fig:global-gate}}
		\vspace{-10pt}
	\caption{Architectures of local and global prosodic gating mechanisms.}
	\label{fig:local}
	\vspace{-10pt}
\end{figure}

High tone/energy sounds can appear in a few segments of the whole input audio. However, these sounds can not contribute equally to the E2E DAC task. 
Inspired by the gating mechanism in the LSTM architecture~\cite{hochreiter1997long}, we extend a local prosodic gating to selectively combine each local prosodic features $\prosodyframe_i$ in Eq.~(\ref{eq:prosody}) with LFBE features $\lfbe_i$ for each audio frame $\audioframe_i$. 
 The local prosodic gating provides a soft mechanism to allow the model to incorporate local prosodic features $\prosodyframe_i$ when needed. Fig. \ref{fig:local}(a) illustrates the architecture of our local prosodic gating, \kai{which operates as follows:}


\noindent\textbf{Input:} A stack $\prosodylist=\{\prosodyframe_1, \prosodyframe_2,...,\prosodyframe_t\}$ of local prosodic features 
and a stack $\lfbelist$ of local LFBE features.

\noindent\textbf{Processing:} A local prosodic gating score $\localgate_i$ is computed from the transformed $\prosodyframe_i$, the transformed $\lfbe_i$, and the interactive features between $\prosodyframe_i$ and $\lfbe_i$. We compute $\localgate_i$ as follows:

\begin{equation}
\label{eq:local}
\localgate_i = \sigma \big(  W^{p}  \prosodyframe_i +  W^{l} \lfbe_i + (W^{lp} \lfbe_i) \otimes  \prosodyframe_i  
\big),
\end{equation}
where $\sigma$ is the \emph{sigmoid} function,  $\otimes$ is the element-wise product operator, $W^{p}, W^{l}$, and $W^{lp}$ are learnable parameters.


%

\noindent\textbf{Output:} A stack $\acousticlist$ of \kai{local acoustic embeddings}, where $\acoustic_i $ is computed as follows:
\vspace{-3pt}
\begin{equation}
\label{eq:fused-local}
\acoustic_i = [\localgate_i \otimes \prosodyframe_i; \lfbe_i]
\end{equation}

\noindent As shown in Eq.~(\ref{eq:local}) and (\ref{eq:fused-local}), when the local prosodic gating score $\localgate_i \rightarrow 1$, $\acoustic_i$ generalizes a simple concatenation between $\prosodyframe_i$ and $\lfbe_i$. In contrast, when $\localgate_i \rightarrow 0$, $\acoustic_i$ simply ignores prosodic signals $\prosodyframe_i$ and only keep $\lfbe_i$. Hence, our local prosodic gating provides a flexible mechanism to effectively fuse $\prosodyframe_i$ with $\lfbe_i$.

\vspace{-10pt}




\subsection{Acoustic encoder}
The acoustic encoder uses the \kai{local acoustic embeddings} to produce global acoustic embeddings. Specifically, its \textbf{inputs} are the stack of fused local acoustic features $\acousticlist$ from the local prosodic gating.
We encode $\acousticlistname$ using a $n$-layer Bi-LSTM acoustic encoder to learn the audio representations. The \textbf{outputs} are a stack $\lstmhiddenlist$ of output hidden states at the last layer $n$ computed as follows:
\begin{equation}
\resizebox{0.9\linewidth}{!}{$
	\begin{aligned}	
		h_i^{(k)} &= W_h^{(k)} \big[\overrightarrow{LSTM}(h_i^{(k-1)}, \overrightarrow{h}_{i-1}^{(k)}  ) ; \overleftarrow{LSTM}(h_i^{(k-1)}, \overleftarrow{h}_{i+1}^{(k)} )\big] 
		\\
		\text{with } i &\in [1, t] \text{, }  
		\overrightarrow{h}_0^{(k)} = \overrightarrow{0} \text{, } 
		\overleftarrow{h}_{t+1}^{(k)} =\overrightarrow{0} \text{, } 
		\text{ and } h_i^{(0)} = \acoustic_i 
	\end{aligned}
$}
	\label{equa:bi-lstm}
\end{equation}
\thanh{where $\overrightarrow{h}_i^{(k)}$ and $\overleftarrow{h}_i^{(k)}$ are the hidden states at time frame $i$ and layer $k$, which are learning from \emph{left-to-right} and  \emph{right-to-left}, respectively.}

\subsection{Global prosodic infusion} \label{prosodySect}
 
The global prosodic gating encodes prosodic features from the entire audio stream and fuses them with the acoustic encoder outputs via our proposed global prosodic gating mechanism (see Fig. 2(b)).
\subsubsection{Global Prosodic Encoder}
Inspired by \cite{Tran_CNN}, we design a 2-D CNN module to capture global prosodic signals at varying timescales using multiple convolution filters. 
The output from each filter is max-pooled, stacked, and flattened, resulting in an output feature matrix $V$. In our experiments, we set the stride to 1 and use four different kernel lengths (5, 10, 25, 50). 
\vspace{-5pt}
\subsubsection{Global Prosodic Gating}
The global prosodic gating selectively fuses the global prosodic features $\globalprosodiclistname$ produced by the CNN encoder and the global acoustic features produced by the acoustic encoder $H$. The most common fusion method is the attention method with the \emph{softmax} operator \cite{vaswani2017attention,bahdanau2014neural}, where  $\lstmhiddenlist$ is the input query and $\globalprosodiclistname$ are the input key and value. However, such methods are not able to \textit{delete} or \textit{subtract} prosodic \kai{embeddings} from the acoustic information. To selectively fuse the global prosodic features, we design a global prosodic gating layer based on \cite{Tay} that learns in parallel a pair-wise similarity matrix and a pair-wise dissimilarity matrix between global \kai{prosodic embeddings} and global acoustic \kai{embeddings produced by the acoustic encoder}. Under this dual affinity scheme, the pair-wise similarity matrix is followed by the \emph{tanh} function, resulting in similarity scores between [-1, 1], which controls the \emph{addition} and \emph{subtraction} of prosodic embeddings from acoustic embeddings. The pair-wise dissimilarity matrix, on the other hand, is served as a gating mechanism that erases prosodic-acoustic similarity  scores to \emph{zero} when prosodic information is not necessary. We detail our proposed global gating below:

\noindent\textbf{Input:} Global prosodic features $\globalprosodiclistname$  and global acoustic features $\lstmhiddenlistname$.

\noindent\textbf{Processing:} 
As shown in Fig. 2(b), we first project each $\lstmhidden_i \in \lstmhiddenlistname$ and $\globalprosodic_j \in \globalprosodiclistname$ into a space with the same dimension. This serves our goal of measuring affinity matrices between $\lstmhiddenlistname$ and $\globalprosodiclistname$.
\begin{equation}
\label{eq:global-project}
\lstmhidden_i ^{\prime} = W^{\lstmhidden} \lstmhidden_i 
\text{,} \;\;\; \globalprosodic_j ^{\prime} = W^{\globalprosodic} \globalprosodic_j 
\end{equation}

\noindent Next, we compute an affinity matrix $\similaritymatrix$, which measures pair-wise similarities between $\lstmhiddenlistname$ and $\globalprosodiclistname$, where each entry $\similaritymatrix_{ij}$ indicates a pair-wise similarity score between $\lstmhidden_i \in \lstmhiddenlistname$ and $\globalprosodic_j \in \globalprosodiclistname$. $\similaritymatrix_{ij}$ is measured as follows:
\vspace{-3pt}
\begin{equation}
	\label{eq:global-sim}
	\similaritymatrix_{ij} = \lstmhidden_i^\prime . {\globalprosodic^\prime}_j^T 
\end{equation}

\noindent Before computing \emph{tanh}($\similaritymatrix$), we want to ensure that $\similaritymatrix$ has both positive and negative values, which encapsulates both the signal addition and subtraction. Thus, we first normalize $\similaritymatrix$ to have a \emph{zero} mean, then applying the \emph{tanh} function on the normalized $\similaritymatrix$.
\vspace{-3pt}
\begin{equation}
	\label{eq:global-sim-norm}
	S = tanh[ \similaritymatrix - mean(\similaritymatrix)]
	\vspace{-3pt}
\end{equation}

\noindent In a same manner, we formulate an affinity matrix $\dissimilaritymatrix$, which measures pair-wise dissimilarities between $\lstmhiddenlistname$ and $\globalprosodiclistname$.  
\vspace{-3pt}
\begin{equation}
	\label{eq:global-dis-sim}
	\dissimilaritymatrix_{ij} = - \Vert \lstmhidden_i^\prime, {\globalprosodic^\prime}_j \Vert_{l_1},
\end{equation}
where $\Vert \cdot \Vert_{l_1}$ indicates the $L_1$ distance between two input feature vectors. From $\dissimilaritymatrix$, we formulate a gating matrix $\globalgatematrix$, which acts as a mechanism to erase unnecessary global prosodic signals by:
\vspace{-3pt}
\begin{equation}
	\label{eq:global-gate}
	\globalgatematrix = \sigma [ \dissimilaritymatrix - mean(\dissimilaritymatrix) ],
\end{equation}
where $\sigma$ is the \textit{sigmoid} function. Since $L_1$ distance is non-negative, $\sigma(\dissimilaritymatrix) \in [0, 0.5]$. Hence, we normalize $\dissimilaritymatrix$ to have a \emph{zero} mean (see Eq. 9)
which ensures $\globalgatematrix \in [0, 1]$.

\noindent\textbf{Output:} We produce a matrix $F$ as the fusion of $\lstmhiddenlistname$ and $\globalprosodiclistname$ by concatenating $\lstmhiddenlistname$  with the attended $\globalprosodiclistname$ as follows:
\begin{equation}
	\label{eq:globalfusion}
	F = [\lstmhiddenlistname; (S \otimes G) \globalprosodiclistname]
\end{equation}

\noindent Last, we apply the \emph{max-pooling} operator on $F$ to obtain a final representation vector $f=\text{\emph{max-pooling}}(F)$ of the input audio $X$ and use it for the DAC task.

\subsection{Dialogue act classification}
For each input audio $X$, we use the acoustic representation vector $f$ as the output of the global prosodic infusion component and produce a DA distribution over all $D$ DAs in the input dataset. The cross entropy loss for the input audio $X$ is defined as:
\begin{equation}
\label{eq:loss}
\begin{aligned}
	\widehat{y}^{diag}_X &= \text{softmax} (W^f f ) 
	\\
	\mathcal{L}_X 			&= -\sum_{d=1}^{D}  {y}^{diag}_{X, d}	\text{log}\big( \widehat{y}^{diag}_{X, d}   \big)
\end{aligned}
\end{equation}

\section{Experiment Settings}
\noindent\textbf{Datasets:}
We use three public benchmark datasets to train and evaluate our proposed models: DSTC2 \cite{second}, DSTC3 \cite{third}, and Switchboard Dialogue Act corpus (SwDA) \cite{godfrey1992switchboard, Jurafsky1998switchboard}. Table~\ref{table:datasets} provides descriptions for each dataset.

The DSTC2 is a standard dataset for tracking the dialogue state. Each utterance has a corresponding audio recording and the associated DAs. The DA is represented in a triple of the following form (actionType, slotName, slotValue). In this work, we treat each utterance in a dialogue as independent because our focus is to examine whether prosodic contexts of the current utterance is useful to its DAC or not. If an utterance only contains one DA label, we use that label. If an utterance contains more than one label, we combine all the labels for that utterance into a single label. In total, the DSTC2 has 15 unique DA labels. 

The DSTC3 is an extension of DSTC2 to a broader domain without providing any further in-domain training data. We adopted the same labelling strategy as DSTC2. In total, the DSTC3 has 17 unique DA labels\footnote[1] {Both DSTC2 and DSTC3 datasets are available at https://github.com/matthen/dstc. }. 

The SwDA is a collection of 1,155 five-minute telephone conversations between 543 speakers of American English. It was originally collected by \cite{godfrey1992switchboard}. The DAs were annotated as a part of the SWBD-DAMSL project \cite{Jurafsky1998switchboard}.  We identified the corresponding audio for each annotated split using the unique conversation id for each utterance. In total, there are 42 unique DA labels\footnote[2]{The annotated DA train, validation, and test splits are available at https://github.com/NathanDuran/Switchboard-Corpus}.


\vspace{-25pt}
  \begin{center}
  \begin{table}[ht!]
  \caption{Number of utterances (hours) of each dataset}
      \resizebox{0.9\linewidth}{!}{
        \begin{tabular}{l|ccc}
         \hline
         & Train & Validation & Test\\         
         \hline
         
         DSTC2 &   12,930 (4.6) & 1,437 (0.5)& 9,116 (3.2)\\

         DSTC3 &  10,870 (8.7) & 1,551 (1.3 ) & 3,100 (2.5) \\
        
          SwDA & 192,768 (289.1) & 3,196 (4.8) & 4,088 (6.7 ) \\
           \hline
        \end{tabular}
        }
        
      \label{table:datasets}
      \vspace{-10pt}
      \end{table}
\end{center}
\noindent\textbf{Baseline models:} \kai{We build an E2E DAC model without any prosodic features (hereafter, baseline), and a model where prosodic information is concatenated with LFBEs at the local level (hereafter, local concat). We also report publicly available E2E DAC accuracy from \cite{ortega2018lexico} and \cite{he2018exploring}. Further, we \kai{evaluate} our proposed encoder against the state-of-the-art prosody encoder \cite{Tran_CNN}.}

\noindent\textbf{Experimental Setup:} \thanh{We report test set accuracy of E2E DAC in all datasets.}
All experiments are implemented \kai{by} using PyTorch \cite{PyTorch}. Training is performed using the Adam optimizer \cite{adam} with $\beta_1 = 0.9$, $\beta_2 =
0.999$, and $\epsilon = 10^{-8}$. The initial learning rate was set to 1e-4 and 5e-4 for the DSTC2 and DSTC3 datasets, respectively. We use a batch size of 32 and train for 60 epochs, with check-pointing based on validation loss. We run each experiment 10 times and report the mean and standard deviation of the accuracy score. The Mann-Whitney U test \cite{mcknight2010mann} is used to determine the statistical significance level of the proposed model accuracy improvement. For the LSTM acoustic encoder, we use a three-layer Bi-LSTM acoustic encoder, with each layer containing 512 hidden units. 


\section{Results}

\noindent\textbf{Overall Accuracy}:  Table~\ref{table:overallResults} shows the overall accuracy of our proposed models and baselines on three benchmark datasets. 
We observe that a simple concatenation method (local concat) that combines LFBEs with prosodic information at frame level improves accuracy by 0.69\% absolute for DSTC3 and 0.4\% absolute for SwDA ($p<0.05$). Further, our proposed model improves E2E DAC across all three benchmark datasets, with 0.39\%, 1.65\%, and 1.17\% absolute increases in accuracy ($p<0.05$) on DSTC2, DSTC3, and SwDA, respectively, \kai{suggesting the critical role of prosodic information in E2E DAC tasks.}  


\vspace{-10pt}
\begin{table}[h!]
	\caption{Overall model accuracy. * indicates a significant increase from the baseline (Mann-Whitney U test, \emph{p} $<$ 0.05)}

	\centering
	\resizebox{0.9\linewidth}{!}{
	\begin{tabular}{l|ccc}
		\hline
																& DSTC2 & DSTC3 & SwDA\\
		\hline
		Baseline 											& 93.18$\pm$.52 & 91.01 $\pm$.39 & 55.80$\pm$.56  \\
		Ortega et al. \cite{ortega2018lexico} & -- & --& 50.9  \\
		He et al. \cite{he2018exploring} 		& -- & --& 56.19   \\
		Local concat 									 & 93.23$\pm$.40 & 91.70$\pm$.51* & 56.20$\pm$.48* \\
		\hline
		Our model & \textbf{93.57}$\pm$.30 	   & \textbf{92.66}$\pm$.30* & \textbf{56.97}$\pm$.46* \\
		\hline
	\end{tabular}
	}
	\label{table:overallResults}
\end{table}

\noindent\textbf{The effects of local and global gating}: Table~\ref{table:gatingResults} shows the effects of our proposed gating method. We individually removed gating at the global level and local level. Overall, we observe that removing gating would lead to performance degradation across three benchmark datasets. For example, removing global gating (our model vs. global encoder) decreases absolute accuracy by 0.15\%, 0.72\%, and 0.22\% on DSTC2, DSTC3, and SwDA, respectively. 
It is worth noting that our proposed prosody encoder method also outperforms the global encoder approach proposed by~\cite{Tran_CNN}.

\vspace{-10pt}
	\begin{table}[h!]
		\caption{Effect of gating mechanisms on E2E DAC accuracy.}
	
		\centering
		\tiny
		\resizebox{0.98\linewidth}{!}{
		\begin{tabular}{l|ccc}
			\hline
			& DSTC2 & DSTC3 & SwDA\\
			\hline
			Our model  & \textbf{93.57}$\pm$.30 & \textbf{92.66}$\pm$.30 & \textbf{56.97}$\pm$.46 \\
			Global encoder~\cite{Tran_CNN} & 93.42$\pm$.33 & 91.94$\pm$.42 & 56.75$\pm$.31 \\
			Local gating & 93.41$\pm$.29 & 91.91$\pm$.33 & 56.71$\pm$.40  \\
			Local concat & 93.23$\pm$.40 & 91.70$\pm$.51 & 56.20$\pm$.48 \\
			\hline
		\end{tabular}
		}
		\label{table:gatingResults}
	\end{table}

\noindent{\textbf{The effects of pitch and energy features:}}~Table~\ref{table:prosodyResults} shows the effects of pitch and energy features when training our best-performing model.  Specifically, we investigate the performance when removing the energy feature group and when removing the pitch feature group.  We observe pitch features are more critical than energy features for our proposed model. For example, the accuracy of our proposed model drops almost 1\% in the absolute E2E accuracy for SwDA when pitch features are absent, whereas  the accuracy of our proposed model drops only 0.6\% when energy features are absent.
\vspace{-11pt}
\begin{table}[ht!]
	\caption{Effect of different prosodic information.}
	\label{table:prosodyResults}
	\centering
	\tiny
	\resizebox{0.92\linewidth}{!}{
	\begin{tabular}{l|ccc}
	\hline
						& DSTC2 & DSTC3 & SwDA  \\
			\hline
	Our model		  & \textbf{93.57}$\pm$.30 & \textbf{92.66}$\pm$.30 & \textbf{56.97}$\pm$.46 \\
	w/o energy   & 93.37$\pm$.12 & 92.53$\pm$.24 & 56.38$\pm$.49 \\
	w/o pitch 	   & 93.30$\pm$.28 & 92.25$\pm$.27 & 55.99$\pm$.27  \\
	    		\hline

	\end{tabular}
}

\end{table}

\vspace{-15pt}
 \begin{figure}[htp]
 \caption{DA label-wise accuracy and distribution of gating scores.}
    \centering
      \includegraphics[width=1\linewidth]{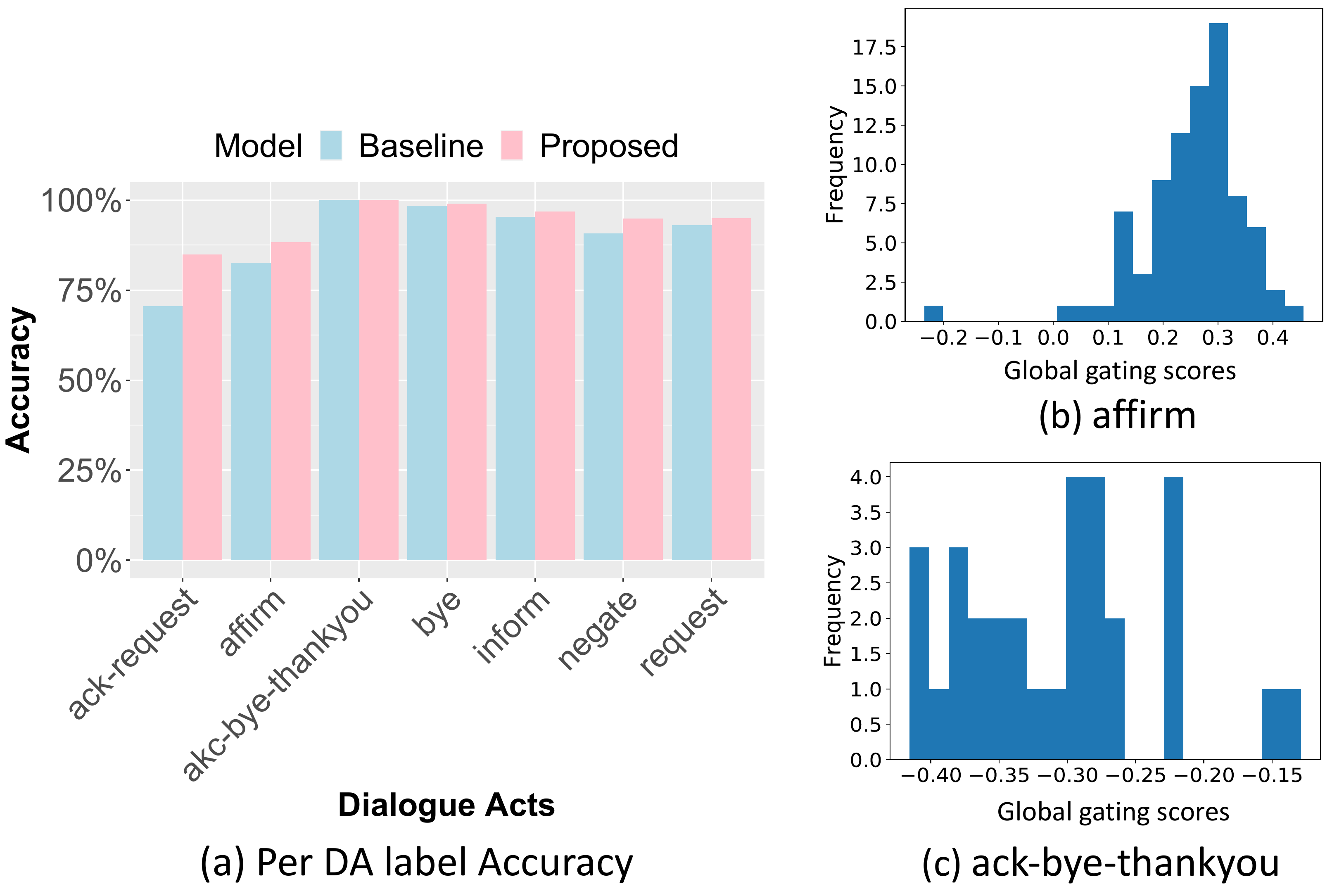}

    \label{fig:per_da}
\end{figure}

\noindent{\textbf{DA label-wise analysis:}} \kai{Fig.~\ref{fig:per_da} shows the DA label-wise accuracy and gating scores for our proposed model on the DSTC3 dataset. The \emph{ack-request} gains the most improvement, with a 14.26\% absolute increase in accuracy (see Fig. 3(a)). After listening to the audio, we found that the \emph{ack} and \emph{request} sequences are connected without any pause, with the \emph{ack} consisting of only one word (e.g., okay). 
This finding is consistent with ~\cite{gravano2007on,benus2007the}'s linguistic observation where affirmative cue words such as \emph{okay} can be distinguished by rising pitch. Our proposed gating mechanism was expected to solve complex situations, such as multiple dialogue acts, with one localized in a short portion of the utterance. To verify, we show the distributions of the global gating scores for \emph{affirm} (the DA class where our model outperforms the base model and also contains mostly affirmative cue words such as \emph{yes}) and \textit{ack-bye-thankyou} (the DA class where our model and the base model perform similarly) in Figure \ref{fig:per_da}(b) and Figure \ref{fig:per_da}(c), respectively. 
The gating scores of \textit{affirm} is mostly distributed in [0.00, 0.40], indicating that prosodic features contribute positively and help improve our model's performance. In contrast, the scores of \textit{ack-bye-thankyou} are mostly distributed in [-0.50, -0.1], suggesting that prosodic features contribute negatively and do not help the performance. This suggests the need of our gating mechanisms \thanh{to} correctly recognize DA labels.}

\section{Conclusion}

In this work, we introduced a novel neural model architecture to integrate prosodic contexts into an acoustic encoder for E2E DAC. We used pitch and energy related features as a measure for prosodic contexts and integrated them at both the local level and global level. Our experiments show that our proposed approach \kai{provides improvements over the state-of-the-art solutions}. In the future, we plan to pave the way for fusing the proposed neural architecture with other prosody-related acoustic cues, such as speaking-rate.

\bibliographystyle{IEEEbib}
\bibliography{mybib}

\end{document}